\pdfoutput=1

\documentclass[11pt]{article}

\usepackage[preprint]{acl}

\usepackage{times}
\usepackage{latexsym}

\usepackage[T1]{fontenc}

\usepackage[utf8]{inputenc}

\usepackage{microtype}

\usepackage{inconsolata}

\usepackage{graphicx}
\usepackage{amsmath}
\usepackage{amssymb}
\usepackage{multirow}
\usepackage{colortbl}
\usepackage{amsmath}
\usepackage[normalem]{ulem}
\usepackage{subcaption}
\useunder{\uline}{\ul}{}

\title{Can Large Language Models Address Open-Target Stance Detection?}

  \author{Abu Ubaida Akash \quad Ahmed Fahmy \quad Amine Trabelsi \\ 
            Department of Computer Science, Université de Sherbrooke \\
            \texttt{\{akaa2803,mosa2801,amine.trabelsi\}@usherbrooke.ca}}

\begin{document}
\maketitle
\begin{abstract}

Stance detection (SD) identifies a text's position towards a target, typically labeled as \textit{favor}, \textit{against}, or \textit{none}. We introduce Open-Target Stance Detection (OTSD), the most realistic task where targets are neither seen during training nor provided as input. We evaluate Large Language Models (LLMs) from GPT, Gemini, Llama, and Mistral families,
comparing their performance to the only existing work, Target-Stance Extraction (TSE), which benefits from predefined targets. Unlike TSE, OTSD removes the dependency of a predefined list, making target generation and evaluation more challenging. We also provide a metric for evaluating target quality that correlates well with human judgment. Our experiments reveal that LLMs outperform TSE in target generation, both when the real target is explicitly and not explicitly mentioned in the text. Similarly, LLMs overall surpass TSE in stance detection for both explicit and non-explicit cases. However, LLMs struggle in both target generation and stance detection when the target is not explicit.\footnote{Dataset and code are available here: \url{https://github.com/AbuUbaida/opentarget}.}
\end{abstract}

\section{Introduction}
\label{sec:intro}


Stance detection (SD) aims to determine the position of a text or person towards a certain \textit{target}, typically categorized as “favor”, “against”, or “none”. 
The \textit{target} can be mentioned explicitly in the text, or sometimes the idea of the \textit{target} can be conveyed indirectly \citep{kuccuk2020stance}.
In Zero-shot Stance Detection (ZSSD), a model predicts stances for targets it has not seen during training, which is crucial since collecting training data for every potential target is impractical \citep{allaway2021adversarial}.
While recent research has focused on ZSSD \citep{zhang2023task, li2023tts, wen2023zero, liang2022zero, zhu2022enhancing, luo2022exploiting, xu2022openstance},
most studies assume that the target is known or manually identified and given as input, a rare scenario in real-world applications where the target is often uncommon, unknown or not explicitly conveyed. Also, target annotation is 
an expensive task in SD \citep{kuccuk2020stance}. 

In this paper, we focus on a different yet challenging task we refer to as \textbf{\texttt{Open-Target Stance Detection (OTSD)}}. In this task, the target is neither seen during training (zero-shot) nor provided as input to the model. In OTSD, one important challenge lies in identifying the target from the text rather than having it provided as input. The target may be mentioned explicitly or not and the stance is predicted with respect to the produced target. It is worth noting that in cross-target stance detection \citep{zhang2020enhancing}, while the target is also unseen during training, it typically belongs to domains similar to those in the training set.
In contrast, in the OTSD task, we mostly consider scenarios where targets from unrelated or different domains may emerge. Both tasks fall under the category of ZSSD.
There has been one notable attempt to partially address this real-world setting of SD.
\citet{li2023new} generate targets from the input text and later detect stance based on these targets. However, in target generation, they map the generated targets to a predefined list of golden targets (details in §\ref{sec:related_work}) to ensure exact wording and facilitate evaluation. They call their approach \textbf{\texttt{TSE}}, which does not fully align with our OTSD setting. In OTSD we assume no input target information is given during the whole process, making it more practical than TSE, which requires a comprehensive list of all possible targets.
Moreover, our experiment reveals a performance gap in \citet{li2023new}'s approach when used on text with explicit or non-explicit target mentions, a factor not addressed by the authors that warrants further investigation in the context of OTSD.
In this work, we present and explore the OTSD task, focusing on its two main steps: Target Generation (TG) and Stance Detection (SD) (Eq. \ref{eq:our_task}). As a zero-shot task for both target and stance, Large Language Models (LLMs) appear well-suited for this challenge. Therefore, we examine the performance of LLMs and compare it to the primary existing work, TSE, which benefits from using a predefined list of targets during the process (Eq. \ref{eq:tse}). 
Our empirical study aims to address the following research questions:
[\textbf{RQ\#1}] How do proprietary and open LLMs—particularly models from GPT, Gemini, Llama, and Mistral families,
—perform in open TG compared to TSE when the real target is explicitly or non-explicitly mentioned in the text,
and how do they compare to each other? [\textbf{RQ\#2}] How do the same LLMs perform compared to TSE in SD (in the context of OTSD) for both explicit and non-explicit cases, and how do they compare to each other? Our contributions consist of introducing the task of OTSD, providing a target evaluation metric,
and conducting experiments answering the research questions.

\section{Target-Stance Extraction (TSE)}
\label{sec:related_work}

In TSE, \citet{li2023new} demonstrate target extraction and stance detection tasks in both in-target and zero-shot settings. For our comparison, we only consider their zero-shot setting. In TSE zero-shot, they first generate targets using keyphrase generation models, then map these targets to a predefined list of golden targets to find the closest match, and finally use this match to detect the stance of the text (Eq. \ref{eq:tse}).
The best match from the keyphrase model, marked as \texttt{TSE-BestGen}, and
the best mapped target (\texttt{TSE-mapped}) is considered as the predicted target. For example, given the text \textit{"Embracing different faiths teaches ... 
diverse beliefs and foster unity."}, TSE primarily generates targets such as \textit{Peace}, \textit{Religious diversity}, \textit{Respect}, \textit{etc.}. From a predefined list of possible targets (\textit{e.g.}, \textit{Face Mask}, \textit{Atheism}, \textit{Donald Trump}, \textit{etc.}), \textit{Atheism} matches most closely with \textit{Religious diversity} and is considered as the final predicted target. In OTSD, we aim to generate \textit{Religious diversity} or a closely related phrase directly, bypassing the predefined list. The stance is then detected toward this generated target.

\section{Open-Target Stance Detection (OTSD)}
\label{sec:open-target-sd}

\subsection{Task Definition}
\label{subsec:task}

Given an input text \(x\), previous work \cite{li2023new} (Eq.~\ref{eq:tse}) generates a target \(t'\) and maps it to a predefined list of targets \(k\), resulting in a mapped target \(t\). Finally, they detect the stance \(y\) given \(x\) and \(t\). OTSD objective (Eq.~\ref{eq:our_task}) is to detect the stance \(y\) from the input text \(x\) and the generated target \(t'\), eliminating the need for a predefined list. 



\setlength{\abovedisplayskip}{0pt} \setlength{\abovedisplayshortskip}{0pt}

\begin{gather}
\label{eq:tse}
x \xrightarrow{\text{generate}} t' \xrightarrow{\text{map}} t \in k, \quad x + t \rightarrow y \\
\label{eq:our_task}
x \xrightarrow{\text{generate}} t', \quad x + t' \rightarrow y
\end{gather}

\subsection{Approach}
\label{subsec:approach}
To address the research questions outlined in §\ref{sec:intro}, our approach to OTSD leverages the zero-shot learning capabilities of LLMs.
We adopt the "Task Definition" prompting strategy, as demonstrated by \citet{cruickshank2023use}, which either outperforms or performs on par with alternative strategies such as "Chain-of-Thought" (CoT) and "few-shot" across most stance detection datasets. To further validate this, we conduct a preliminary experiment with the CoT strategy (detailed in Appendix \ref{app:cot_init_result}), which reveals that CoT underperforms in TG, and performs comparably to "Task Definition" in SD, reinforcing the early
findings. Additionally, using the "Task Definition" strategy without relying on tailored prompts for specific targets or in-context examples ensures the generalizability of OTSD, aligning with the zero-shot nature of this task. This approach first names the task and then provides a concise, self-contained definition of what that task entails—its purpose, inputs, and expected outputs—before asking the model to perform it.
We employ two different prompting approaches for TG and SD. The prompts and their design justifications can be found in Appendix \ref{app:prompt}.
\textbf{TG+SD (Two-Step Approach)} In this approach, we first generate the target and then sequentially detect the stance of the text towards the generated target,
which directs LLMs to focus on one task at a time.
\textbf{TG\&SD (Single-Step Joint Approach)} 
This approach unifies TG and SD in a single step using a prompt, enabling the model to better understand the relationship between the text, target, and stance.



\section{Experiments}


In OTSD task, we compare our approaches with TSE in both TG and SD. To ensure sensitivity and robustness, we run our experiments three times independently and report the average results \footnote{Score variations are negligible, up to four decimal places.}. Since OTSD is a generative task, we primarily compare our results with \texttt{TSE-BestGen}, despite its reliance on a predefined list of golden targets.
Details about our hardware settings can be found in Appendix \ref{app:hardware_settings}.

\begin{table}[t]
\centering
\setlength{\tabcolsep}{2.5 pt} 
\resizebox{\linewidth}{!}
{%
\begin{tabular}{lccccc}
\hline
\multicolumn{1}{c}{\multirow{2}{*}{\textbf{Dataset}}} & \multirow{2}{*}{\textbf{Source}} & \multicolumn{2}{c}{\textbf{\#samples}} & \multirow{2}{*}{\textbf{\begin{tabular}[c]{@{}c@{}}\#unique\\ -targets\end{tabular}}} & \multirow{2}{*}{\textbf{\#stance}} \\ \cline{3-4}
\multicolumn{1}{c}{} &  & \multicolumn{1}{l}{\textbf{Explicit}} & \multicolumn{1}{l}{\textbf{Non-explicit}} &  &  \\ \hline
TSE & Tweets & 1,804 & 1,196 & 6 & 3 \\
VAST & News comments & 3,120 & 1,980 & 2,145 & 3 \\
EZSTANCE & Tweets & 9,313 & 149 & 6,873 & 3 \\ \hline
\end{tabular}%
}
\caption{Statistical overview of the datasets utilized for the OTSD task.}
\label{tbl:dataset_stat}
\end{table}

\subsection{Dataset and Model}
\label{subsec:dataset_model}
To ensure a fair comparison between TSE and LLM models, we use the same dataset as TSE’s zero-shot setup. In addition to TSE dataset, we include VAST \citep{allaway-mckeown-2020-zero} and EZSTANCE \citep{zhao-caragea-2024-ez} to evaluate LLMs across different settings. As shown in Table \ref{tbl:dataset_stat}, all three datasets contain three stance classes (\textit{Favor}, \textit{Against}, \textit{None}) and include 3000, 5100, and 9313 samples, respectively, in our study.
To align with the scope of our study, the multi-target VAST and EZSTANCE datasets are converted into single-target formats (details in Appendix \ref{app:dataset_proc}). TSE, VAST, and EZSTANCE feature 6, 2145, and 6873 unique targets, respectively. These datasets—particularly VAST and EZSTANCE—cover a wide range of domains and unique targets that are rarely used in stance detection, highlighting the practical challenges of the OTSD task.
Samples are classified as explicit or non-explicit through preprocessing steps, including stop-word and special character removal, word lemmatization, and checking for target words in the text. This process results in 1804 explicit and 1196 non-explicit samples from TSE dataset, 3120 explicit and 1980 non-explicit samples from VAST (see Appendix \ref{app:dataset_samples}), and 9313 explicit and 149 non-explicit samples from EZSTANCE.
Note that due to the lack of code and details from the TSE method for reproducing target generation and selection, we could not use other SD datasets for benchmarking.
For models, we use both proprietary and non-proprietary LLMs, including GPT-3.5, GPT-4o, Gemini-flash-8B, Gemini-pro, Llama-3-8B, Llama-3-70B, Mistral-small, and Mistral-large (description in Appendix ~\ref{app:model_description}).

\begin{table*}[t]
\centering
\setlength{\tabcolsep}{2.5 pt} 
\resizebox{\textwidth}{!}{%
\begin{tabular}{l|c|cccccccc|c|cccccccc|c|cccccccc}
\hline
\multicolumn{1}{c|}{} &  & \multicolumn{4}{c|}{TG+SD} & \multicolumn{4}{c|}{TG\&SD} &  & \multicolumn{4}{c|}{TG+SD} & \multicolumn{4}{c|}{TG\&SD} &  & \multicolumn{4}{c|}{TG+SD} & \multicolumn{4}{c}{TG\&SD} \\
\multicolumn{1}{c|}{\multirow{-2}{*}{Model}} &  & SS & BTSD & HE & \multicolumn{1}{c|}{SC} & SS & BTSD & HE & SC &  & SS & BTSD & HE & \multicolumn{1}{c|}{SC} & SS & BTSD & HE & SC &  & SS & BTSD & HE & \multicolumn{1}{c|}{SC} & SS & BTSD & HE & SC \\ \cline{1-1} \cline{3-10} \cline{12-19} \cline{21-28} 
 &  & \multicolumn{8}{c|}{\cellcolor[HTML]{D5D5D5}Explicit} &  & \multicolumn{8}{c|}{\cellcolor[HTML]{D5D5D5}Explicit} &  & \multicolumn{8}{c}{\cellcolor[HTML]{D5D5D5}Explicit} \\
TSE-M &  & {\ul 0.96} & 36.63 & \multicolumn{1}{c|}{{\ul 0.716}} & \multicolumn{1}{c|}{38.10} & - & - & \multicolumn{1}{c|}{-} & - &  & - & - & \multicolumn{1}{c|}{-} & \multicolumn{1}{c|}{-} & - & - & \multicolumn{1}{c|}{-} & - &  & - & - & \multicolumn{1}{c|}{-} & \multicolumn{1}{c|}{-} & - & - & \multicolumn{1}{c|}{-} & - \\
TSE-B &  & 0.86 & 35.8 & \multicolumn{1}{c|}{0.338} & \multicolumn{1}{c|}{37.81} & - & - & \multicolumn{1}{c|}{-} & - &  & - & - & \multicolumn{1}{c|}{-} & \multicolumn{1}{c|}{-} & - & - & \multicolumn{1}{c|}{-} & - &  & - & - & \multicolumn{1}{c|}{-} & \multicolumn{1}{c|}{-} & - & - & \multicolumn{1}{c|}{-} & - \\
GPT-3.5 &  & 0.87 & 38.43 & \multicolumn{1}{c|}{0.546} & \multicolumn{1}{c|}{42.68} & 0.87 & \textbf{39.60} & \multicolumn{1}{c|}{\textbf{0.663}} & \textbf{47.61} &  & 0.82 & 41.67 & \multicolumn{1}{c|}{0.461} & \multicolumn{1}{c|}{47.21} & \textbf{0.89} & \textbf{44.25} & \multicolumn{1}{c|}{\textbf{0.581}} & \textbf{48.48} &  & 0.83 & 49.27 & \multicolumn{1}{c|}{0.431} & \multicolumn{1}{c|}{39.10} & {\ul \textbf{0.88}} & \textbf{49.70} & \multicolumn{1}{c|}{\textbf{0.439}} & \textbf{40.63} \\
GPT-4o &  & 0.87 & {\ul 41.55} & \multicolumn{1}{c|}{0.566} & \multicolumn{1}{c|}{44.78} & \textbf{0.88} & {\ul \textbf{41.92}} & \multicolumn{1}{c|}{{\ul \textbf{0.690}}} & \textbf{46.83} &  & 0.84 & {\ul 42.69} & \multicolumn{1}{c|}{0.519} & \multicolumn{1}{c|}{40.77} & \textbf{0.88} & {\ul \textbf{44.25}} & \multicolumn{1}{c|}{\textbf{0.629}} & \textbf{49.38} &  & 0.83 & {\ul 49.70} & \multicolumn{1}{c|}{0.449} & \multicolumn{1}{c|}{45.93} & \textbf{0.87} & {\ul \textbf{50.69}} & \multicolumn{1}{c|}{\textbf{0.469}} & \textbf{46.22} \\
Gemini-flash &  & 0.87 & 40.02 & \multicolumn{1}{c|}{0.593} & \multicolumn{1}{c|}{46.76} & \textbf{0.88} & \textbf{40.87} & \multicolumn{1}{c|}{\textbf{0.624}} & \textbf{47.70} &  & 0.85 & 42.03 & \multicolumn{1}{c|}{0.681} & \multicolumn{1}{c|}{47.31} & \textbf{0.88} & \textbf{42.14} & \multicolumn{1}{c|}{\textbf{0.706}} & \textbf{48.89} &  & 0.84 & 49.33 & \multicolumn{1}{c|}{{\ul 0.482}} & \multicolumn{1}{c|}{46.35} & \textbf{0.87} & \textbf{50.10} & \multicolumn{1}{c|}{{\ul 0.588}} & \textbf{47.26} \\
Gemini-pro &  & 0.86 & 40.81 & \multicolumn{1}{c|}{0.643} & \multicolumn{1}{c|}{40.76} & {\ul \textbf{0.89}} & \textbf{40.92} & \multicolumn{1}{c|}{\textbf{0.664}} & \textbf{45.71} &  & 0.83 & 42.52 & \multicolumn{1}{c|}{{\ul 0.686}} & \multicolumn{1}{c|}{43.64} & \textbf{0.87} & \textbf{42.78} & \multicolumn{1}{c|}{{\ul \textbf{0.707}}} & {\ul \textbf{51.46}} &  & 0.82 & 49.12 & \multicolumn{1}{c|}{0.472} & \multicolumn{1}{c|}{45.53} & \textbf{0.86} & \textbf{50.08} & \multicolumn{1}{c|}{0.521} & \textbf{48.47} \\
Llama-3-8B &  & \textbf{0.88} & \textbf{38.31} & \multicolumn{1}{c|}{0.54} & \multicolumn{1}{c|}{\textbf{43.13}} & 0.87 & 37.75 & \multicolumn{1}{c|}{\textbf{0.558}} & 42.10 &  & 0.86 & 41.13 & \multicolumn{1}{c|}{\textbf{0.467}} & \multicolumn{1}{c|}{\textbf{47.20}} & \textbf{0.88} & \textbf{42.45} & \multicolumn{1}{c|}{0.458} & 46.72 &  & 0.84 & 47.07 & \multicolumn{1}{c|}{0.435} & \multicolumn{1}{c|}{40.51} & \textbf{0.87} & \textbf{48.79} & \multicolumn{1}{c|}{\textbf{0.441}} & \textbf{43.42} \\
Llama-3-70B &  & 0.88 & 41.01 & \multicolumn{1}{c|}{0.570} & \multicolumn{1}{c|}{{\ul 47.18}} & 0.88 & \textbf{41.52} & \multicolumn{1}{c|}{\textbf{0.635}} & {\ul \textbf{49.84}} &  & 0.86 & 40.73 & \multicolumn{1}{c|}{0.652} & \multicolumn{1}{c|}{{\ul 46.39}} & \textbf{0.88} & {\ul \textbf{42.50}} & \multicolumn{1}{c|}{\textbf{0.655}} & \textbf{48.73} &  & {\ul 0.85} & 46.01 & \multicolumn{1}{c|}{0.442} & \multicolumn{1}{c|}{41.64} & \textbf{0.87} & \textbf{48.79} & \multicolumn{1}{c|}{0.442} & \textbf{42.78} \\
Mistral-small &  & 0.85 & 38.47 & \multicolumn{1}{c|}{0.586} & \multicolumn{1}{c|}{46.15} & \textbf{0.87} & \textbf{39.02} & \multicolumn{1}{c|}{\textbf{0.650}} & \textbf{46.32} &  & 0.84 & 42.13 & \multicolumn{1}{c|}{0.490} & \multicolumn{1}{c|}{45.06} & \textbf{0.86} & \textbf{43.71} & \multicolumn{1}{c|}{\textbf{0.526}} & \textbf{46.77} &  & 0.83 & 49.53 & \multicolumn{1}{c|}{0.436} & \multicolumn{1}{c|}{44.49} & \textbf{0.86} & \textbf{50.17} & \multicolumn{1}{c|}{{\ul \textbf{0.479}}} & \textbf{48.72} \\
Mistral-large &  & 0.88 & 39.32 & \multicolumn{1}{c|}{0.543} & \multicolumn{1}{c|}{47.16} & 0.88 & \textbf{41.39} & \multicolumn{1}{c|}{\textbf{0.647}} & \textbf{49.76} &  & 0.86 & 41.71 & \multicolumn{1}{c|}{0.652} & \multicolumn{1}{c|}{46.20} & {\ul \textbf{0.89}} & \textbf{43.13} & \multicolumn{1}{c|}{\textbf{0.671}} & \textbf{51.30} &  & 0.81 & 48.51 & \multicolumn{1}{c|}{0.444} & \multicolumn{1}{c|}{{\ul 47.12}} & \textbf{0.84} & \textbf{50.27} & \multicolumn{1}{c|}{0.475} & {\ul \textbf{50.35}} \\
 &  & \multicolumn{8}{c|}{\cellcolor[HTML]{D5D5D5}Non-explicit} &  & \multicolumn{8}{c|}{\cellcolor[HTML]{D5D5D5}Non-explicit} &  & \multicolumn{8}{c}{\cellcolor[HTML]{D5D5D5}Non-explicit} \\
TSE-M &  & {\ul 0.90} & 30.56 & \multicolumn{1}{c|}{0.391} & \multicolumn{1}{c|}{32.00} & - & - & \multicolumn{1}{c|}{-} & - &  & - & - & \multicolumn{1}{c|}{-} & \multicolumn{1}{c|}{-} & - & - & \multicolumn{1}{c|}{-} & - &  & - & - & \multicolumn{1}{c|}{-} & \multicolumn{1}{c|}{-} & - & - & \multicolumn{1}{c|}{-} & - \\
TSE-B &  & 0.82 & 29.32 & \multicolumn{1}{c|}{0.215} & \multicolumn{1}{c|}{31.00} & - & - & \multicolumn{1}{c|}{-} & - &  & - & - & \multicolumn{1}{c|}{-} & \multicolumn{1}{c|}{-} & - & - & \multicolumn{1}{c|}{-} & - &  & - & - & \multicolumn{1}{c|}{-} & \multicolumn{1}{c|}{-} & - & - & \multicolumn{1}{c|}{-} & - \\
GPT-3.5 &  & \textbf{0.84} & \textbf{33.10} & \multicolumn{1}{c|}{\textbf{0.419}} & \multicolumn{1}{c|}{32.66} & 0.83 & 31.32 & \multicolumn{1}{c|}{0.405} & \textbf{33.94} &  & 0.79 & 38.10 & \multicolumn{1}{c|}{0.310} & \multicolumn{1}{c|}{45.54} & {\ul \textbf{0.85}} & \textbf{38.55} & \multicolumn{1}{c|}{\textbf{0.416}} & \textbf{45.80} &  & 0.76 & 38.78 & \multicolumn{1}{c|}{0.377} & \multicolumn{1}{c|}{36.69} & \textbf{0.81} & \textbf{41.89} & \multicolumn{1}{c|}{\textbf{0.402}} & \textbf{38.97} \\
GPT-4o &  & \textbf{0.84} & 35.14 & \multicolumn{1}{c|}{0.430} & \multicolumn{1}{c|}{{\ul 36.39}} & 0.83 & {\ul \textbf{36.12}} & \multicolumn{1}{c|}{{\ul \textbf{0.513}}} & \textbf{37.50} &  & 0.80 & 38.55 & \multicolumn{1}{c|}{0.387} & \multicolumn{1}{c|}{39.92} & \textbf{0.83} & \textbf{39.84} & \multicolumn{1}{c|}{\textbf{0.450}} & \textbf{43.84} &  & 0.77 & 42.78 & \multicolumn{1}{c|}{0.410} & \multicolumn{1}{c|}{38.50} & \textbf{0.80} & \textbf{45.70} & \multicolumn{1}{c|}{\textbf{0.436}} & \textbf{39.26} \\
Gemini-flash &  & \textbf{0.84} & 34.59 & \multicolumn{1}{c|}{0.347} & \multicolumn{1}{c|}{34.32} & 0.83 & \textbf{34.73} & \multicolumn{1}{c|}{\textbf{0.365}} & \textbf{35.62} &  & 0.81 & 39.22 & \multicolumn{1}{c|}{0.306} & \multicolumn{1}{c|}{44.87} & \textbf{0.83} & \textbf{39.27} & \multicolumn{1}{c|}{\textbf{0.394}} & \textbf{45.15} &  & 0.78 & 41.38 & \multicolumn{1}{c|}{{\ul 0.506}} & \multicolumn{1}{c|}{{\ul 43.48}} & \textbf{0.81} & \textbf{45.79} & \multicolumn{1}{c|}{{\ul 0.536}} & \textbf{44.48} \\
Gemini-pro &  & 0.83 & 33.44 & \multicolumn{1}{c|}{0.365} & \multicolumn{1}{c|}{32.58} & \textbf{0.84} & \textbf{34.85} & \multicolumn{1}{c|}{\textbf{0.453}} & \textbf{35.96} &  & 0.80 & 39.90 & \multicolumn{1}{c|}{{\ul 0.447}} & \multicolumn{1}{c|}{41.39} & \textbf{0.82} & \textbf{40.53} & \multicolumn{1}{c|}{{\ul \textbf{0.482}}} & {\ul \textbf{48.53}} &  & 0.76 & 40.30 & \multicolumn{1}{c|}{0.482} & \multicolumn{1}{c|}{41.49} & \textbf{0.79} & \textbf{41.01} & \multicolumn{1}{c|}{\textbf{0.518}} & {\ul \textbf{47.36}} \\
Llama-3-8B &  & \textbf{0.85} & 33.06 & \multicolumn{1}{c|}{0.447} & \multicolumn{1}{c|}{\textbf{33.36}} & 0.83 & \textbf{33.81} & \multicolumn{1}{c|}{\textbf{0.490}} & 31.90 &  & 0.82 & 38.74 & \multicolumn{1}{c|}{0.330} & \multicolumn{1}{c|}{{\ul \textbf{47.98}}} & \textbf{0.84} & \textbf{39.91} & \multicolumn{1}{c|}{\textbf{0.383}} & 45.48 &  & 0.78 & 39.68 & \multicolumn{1}{c|}{0.374} & \multicolumn{1}{c|}{\textbf{38.35}} & {\ul \textbf{0.82}} & \textbf{39.75} & \multicolumn{1}{c|}{\textbf{0.386}} & 37.18 \\
Llama-3-70B &  & 0.84 & 32.98 & \multicolumn{1}{c|}{{\ul 0.480}} & \multicolumn{1}{c|}{32.51} & {\ul 0.85} & \textbf{34.67} & \multicolumn{1}{c|}{\textbf{0.488}} & {\ul \textbf{35.50}} &  & {\ul 0.83} & {\ul 40.47} & \multicolumn{1}{c|}{{\ul 0.447}} & \multicolumn{1}{c|}{45.70} & \textbf{0.84} & {\ul \textbf{42.02}} & \multicolumn{1}{c|}{\textbf{0.451}} & \textbf{46.57} &  & 0.77 & 40.15 & \multicolumn{1}{c|}{0.435} & \multicolumn{1}{c|}{42.40} & \textbf{0.80} & \textbf{41.26} & \multicolumn{1}{c|}{0.512} & \textbf{42.93} \\
Mistral-small &  & 0.83 & 30.18 & \multicolumn{1}{c|}{0.386} & \multicolumn{1}{c|}{\textbf{33.61}} & 0.83 & \textbf{34.12} & \multicolumn{1}{c|}{\textbf{0.495}} & 30.47 &  & 0.80 & 39.75 & \multicolumn{1}{c|}{0.405} & \multicolumn{1}{c|}{\textbf{43.34}} & \textbf{0.81} & \textbf{40.05} & \multicolumn{1}{c|}{\textbf{0.447}} & 41.87 &  & 0.77 & 38.71 & \multicolumn{1}{c|}{0.400} & \multicolumn{1}{c|}{\textbf{41.36}} & \textbf{0.78} & \textbf{46.63} & \multicolumn{1}{c|}{\textbf{0.414}} & 39.41 \\
Mistral-large & \multirow{-24}{*}{\rotatebox{90}{TSE}} & \textbf{0.85} & {\ul 35.38} & \multicolumn{1}{c|}{0.465} & \multicolumn{1}{c|}{33.76} & 0.83 & \textbf{35.42} & \multicolumn{1}{c|}{\textbf{0.488}} & \textbf{34.70} & \multirow{-24}{*}{\rotatebox{90}{VAST}} & 0.82 & 39.32 & \multicolumn{1}{c|}{0.394} & \multicolumn{1}{c|}{43.15} & \textbf{0.84} & \textbf{39.98} & \multicolumn{1}{c|}{\textbf{0.424}} & \textbf{46.55} & \multirow{-24}{*}{\rotatebox{90}{EZSTANCE}} & {\ul 0.79} & {\ul 46.65} & \multicolumn{1}{c|}{0.482} & \multicolumn{1}{c|}{39.30} & {\ul \textbf{0.82}} & {\ul \textbf{49.07}} & \multicolumn{1}{c|}{\textbf{0.488}} & \textbf{39.86} \\ \hline
\end{tabular}%
}
\caption{\label{tbl:tg_sd}
Comparison of TG and SD performance between TSE and LLMs using various metrics: TG is evaluated with SS (SemSim), BTSD (\%), and HE (Human Evaluation) scores, while SC (\%) measures SD performance. TSE-M and TSE-B represent \texttt{TSE-mapped} and \texttt{TSE-BestGen}, respectively. Metric-wise best results 
within a specific setting
are \underline{underlined}, and better results of a model across 
TG+SD and TG\&SD
for each dataset
are 
in \textbf{bold}.
}
\end{table*}

\subsection{Evaluation Methods}
\label{subsec:eval_methods}


\textbf{BTSD:}\label{subsec:btsd} To measure the generated target quality, we fine-tune the BERTweet model \citep{nguyen2020bertweet} following the same setup as TSE. We consider 4 datasets combined i.e. SemEval \citep{mohammad2016semeval}, AM \citep{stab2018cross}, COVID-19 \citep{glandt2021stance}, P-Stance \citep{li2021multi}, containing 19 targets, splitting samples by (70\%-train, 10\%-dev, 20\%-test), maintaining equal distribution from each of the targets. To assess the quality of the generated targets, we train a stance classifier model (referred as BTSD) as follows. Given a text $x$ and a target $t$, we first formulate the input as a sequence $s = [[CLS] t[SEP] x]$ where $[CLS]$ is a token that encodes the sentence and $[SEP]$ is used to separate the sentence $x$ and the target $t$. Then the representation of $[CLS]$ token is used to predict the stance toward target $t$. Note that $t$ is the golden target here. During the test phase, initially we considered three random seeds, but finally reported the scores with a single seed, as the score variance was low.

The SD score (F1-macro) is used as the evaluator for the generated targets. To assess the effectiveness of this metric, we experiment with different levels of target quality by modifying or removing words from the golden targets, substituting them with incorrect targets, and selecting random words from the vocabulary as targets. We find that the BERTweet F1 score consistently and significantly improved with higher-quality targets, in both explicit and non-explicit cases (see Appendix \ref{app:BTSD}). More importantly, BTSD score shows strong tau ($\tau$) correlation \citep{kendall1938new} of $0.85$, $0.74$, and $0.79$ for TSE, VAST, and EZSTANCE respectively,
with human judgments gathered during experimentation.
Therefore, BTSD serves as a reliable proxy for evaluating target quality.

\textbf{Human Evaluation:} To evaluate the relevance of the generated targets by some models and further validate the BTSD metric, we conducted a small-scale human evaluation on some of the models using 500 randomly selected samples from each dataset, maintaining an equal ratio of the stance labels. Three annotators were asked to assess the relevance of the generated targets to the gold ones, classifying them as either 0 (Not Related), 0.5 (Partially Related), or 1 (Completely Related) 
(details in Appendix \ref{app:human_ann}). 
We observed a solid overall annotator agreement of $\alpha$=0.76 \citep{krippendorff2011computing}  and $\kappa$=0.664 \citep{fleiss1971measuring}(Appendix Table \ref{tbl:human_eval_corr}).

\textbf{SemSim:} In OTSD, since we do not rely on a predefined list of targets, the generated targets may not precisely match the gold targets but are semantically related. Therefore, we measure the semantic similarity between the generated and gold targets using BERT embeddings. Detailed steps can be found in Appendix \ref{app:semsim} with human judgement correlation of $\tau$=$0.57$ for TSE, and $\tau$=$0.59$ for VAST, way below BTSD.

\noindent We use macro-average F1 for the task of SD.



\section{Result Analysis}
\label{sec:result_analysis}

\subsection{Target Generation (TG)}
\label{subsec:tg_result_analysis}

As shown in Table \ref{tbl:tg_sd}, both human evaluation (HE) and BTSD indicate that LLMs generate higher-quality targets in explicit cases compared to \texttt{TSE-BestGen}, the best model for generating targets not matched to a predefined list. Although \texttt{TSE-mapped} achieves higher HE score in explicit case, its BTSD score is lower. This is because \texttt{TSE-mapped} produces either a perfect match (score=1) or a completely unrelated target (score=0) favouring higher HE score (see Appendix \ref{app:human_score_dist}). In contrast, LLMs mostly generate targets that are highly or partially related to gold targets, resulting in fewer irrelevant outputs and higher BTSD, despite slightly lower average HE scores. 

When the target is not explicitly mentioned, LLMs produce higher-quality targets than both TSE models. However, the quality drops for LLMs (TSE as well) compared to explicit cases (as per Table \ref{tbl:tg_sd}), with HE scores slightly lower than $0.5$ (refers to partially related target) on average. A manual inspection of the lower quality targets suggest that the lower score in non-explicit cases is due to insufficient context of the text. The lack of implicit hints or surrounding context leads the model to generate targets based solely on the text provided, causing inaccuracies. For instance, for the post \textit{"I mean, this reads to me like anyone who's an alcoholic doesn't go to heaven..."}, GPT-3.5 generated the target \textit{"Sinful behavior and heaven's eligibility"} instead of the correct gold target \textit{"gay rights"}.

When comparing tested proprietary and open LLMs using BTSD and HE, there is no clear advantage in TG (\textit{i.e.}, not all proprietary models outperform open ones, or vice versa). Overall, GPT-4o produces targets of equal or higher quality than other tested proprietary models across the approaches, including explicit and non-explicit target mentions. Among the tested open LLMs, Mistral-large and Llama-3-70B outperform each other under different settings.

Comparing between approaches, most models perform better in TG\&SD across both datasets, for explicit and non-explicit cases.

\subsection{Stance Detection (SD)}
\label{subsec:sd_result_analysis}

As shown in Table \ref{tbl:tg_sd}, according to the SC (stance classification), LLMs outperform both TSE models in detecting stances in both explicit and non-explicit cases. In general, LLMs perform better in SD when the target is explicitly mentioned in the text compared to when it is not. Our manual analysis shows that in non-explicit cases, the input text lacks sufficient context about the target, leading to lower SC scores.

The tested proprietary and open LLMs  exhibit competitive performance against each other in both explicit and non-explicit cases for both approaches, with neither showing a clear advantage.

When comparing approaches, LLMs generally perform better with the holistic TG\&SD strategy in both explicit and non-explicit cases, with the exception of Llama-3-8B, and Mistral-small (for non-explicit).

The quality of the generated target (BTSD) shows a stronger correlation with the SC score in explicit cases (avg. $\tau =0.4$) across datasets compared to non-explicit ones (avg. $\tau =0.25$), particularly for VAST and EZSTANCE. 

Upon manual investigation, we observed that LLMs such GPT-4o may generate highly relevant targets that are closely related to the gold targets but nearly `antithetical' or opposite, leading to a stance reversal in comparison to the gold stance. For example, in VAST, when the gold target is \textit{"permit to carry gun"}, GPT-4o generates \textit{"gun control"}, while GPT-3.5 generates \textit{"guns"}. Although GPT-4o's target is more closely related to the gold target, it conveys the opposite. GPT-3.5's target, though broader, aligns with the stance of gold target. Another example, in the explicit sample text, \textit{"Q. from audience: Malcom X said the most disrespected person in America is the black woman, has \#MeToo changed this? Gender has been constructed in the image of the white woman, the idea of a level playing field, white women are often in better positions, says @HeidiMirza \#CHevents"}, GPT-3.5 generated the target \textit{"Gender roles"}, which, while related, fails to capture the specific stance-related target \textit{"MeToo movement"}. This mistake reflects the model's need for broader context to establish the correct link. A potential way to address this issue would be leveraging an LLM as a judge to assess the stance alignment. We explore the feasibility of this idea by conducting a small-scale experiment in Appendix \ref{app:pilot_study_coherence}.
\section{Conclusion}
We introduce the Open Target Stance Detection (OTSD) task, where targets are unseen during training and not provided as input, making it both realistic and challenging. We evaluate proprietary and open LLMs on OTSD and propose a metric for assessing Target Generation (TG) quality. Compared to TSE, which relies on predefined target lists, LLMs perform well in TG but struggle when targets are not explicitly mentioned. LLMs also surpass TSE in Stance Detection (SD) for both explicit and non-explicit targets. Finally, a single-step joint prompt (TG\&SD) outperforms the two-step approach (TG+SD) for both TG and SD.

\section*{Limitation}

The limitations of our work can be listed as follows:

\begin{itemize}
  \item This work has highlighted specific challenges of the proposed OTSD task and measured both target quality and stance detection performance. However, our findings underscore the need to improve the measure of stance detection accuracy in OTSD, enabling it to account for the alignment/coherence between the generated target and the stance toward it, given the text and the reference (gold) target. To investigate the feasibility of a potential solution, we carried out a preliminary experiment using an LLM as an automated judge of stance-target coherence. The LLM's assessments showed strong correlation with human judgments, suggesting that LLMs can serve as a reliable component in the development of coherence-based evaluation metrics. Full details of this pilot study are provided in Appendix \ref{app:pilot_study_coherence}.

  \item 
  In this work, we focus on identifying a single target from each text, even though a given text may contain multiple targets, whether explicitly stated or implied.
  
  \item
  A potential limitation of our work lies in the possibility of data leakage during the pre-training of LLMs, where the models may have been exposed to some of the test data, thereby acquiring prior knowledge of certain entities. Nevertheless, as the LLMs were not explicitly pre-trained on the same task, we maintain that our approach qualifies as zero-shot stance detection.
  
  \item In this work, we limit our experiments to the OTSD task in English. Future research can explore the OTSD task in a cross-lingual setting.
  
  \item Although our findings were evaluated across a diverse set of eight LLMs spanning four different families—covering both small and large models, as well as proprietary and open-source ones—future work could benefit from exploring a broader range of models, particularly Large Reasoning Models (LRMs), which present an especially promising direction.
\end{itemize}

\section*{Ethics Statement}
The datasets used for fine-tuning and zero-shot experiments are publicly available and anonymized, with no associated data protection concerns. Additionally, the human annotators involved in the evaluation were informed about the intended use of their contributions and participated voluntarily without compensation. The only data collected consisted of their annotation ratings.

\section*{Acknowledgments}

This research was partially supported by the Natural Sciences and Engineering Research Council of Canada (NSERC) under the Grant No. RGPIN-2022-04789, and by the Digital Research Alliance of Canada (\url{https://www.alliancecan.ca/en}). We would like to thank the annotators and the reviewers for their valuable feedback.

\bibliography{custom}

\begin{thebibliography}{34}
\providecommand{\natexlab}[1]{#1}

\bibitem[{Achiam et~al.(2023)Achiam, Adler, Agarwal, Ahmad, Akkaya, Aleman, Almeida, Altenschmidt, Altman, Anadkat et~al.}]{achiam2023gpt}
Josh Achiam, Steven Adler, Sandhini Agarwal, Lama Ahmad, Ilge Akkaya, Florencia~Leoni Aleman, Diogo Almeida, Janko Altenschmidt, Sam Altman, Shyamal Anadkat, et~al. 2023.
\newblock Gpt-4 technical report.
\newblock \emph{arXiv preprint arXiv:2303.08774}.

\bibitem[{Allaway and McKeown(2020)}]{allaway-mckeown-2020-zero}
Emily Allaway and Kathleen McKeown. 2020.
\newblock \href {https://doi.org/10.18653/v1/2020.emnlp-main.717} {{Z}ero-{S}hot {S}tance {D}etection: {A} {D}ataset and {M}odel using {G}eneralized {T}opic {R}epresentations}.
\newblock In \emph{Proceedings of the 2020 Conference on Empirical Methods in Natural Language Processing (EMNLP)}, pages 8913--8931, Online. Association for Computational Linguistics.

\bibitem[{Allaway et~al.(2021)Allaway, Srikanth, and McKeown}]{allaway2021adversarial}
Emily Allaway, Malavika Srikanth, and Kathleen McKeown. 2021.
\newblock Adversarial learning for zero-shot stance detection on social media.
\newblock \emph{arXiv preprint arXiv:2105.06603}.

\bibitem[{Brown et~al.(2020)Brown, Mann, Ryder, Subbiah, Kaplan, Dhariwal, Neelakantan, Shyam, Sastry, Askell et~al.}]{brown2020language}
Tom Brown, Benjamin Mann, Nick Ryder, Melanie Subbiah, Jared~D Kaplan, Prafulla Dhariwal, Arvind Neelakantan, Pranav Shyam, Girish Sastry, Amanda Askell, et~al. 2020.
\newblock Language models are few-shot learners.
\newblock \emph{Advances in neural information processing systems}, 33:1877--1901.

\bibitem[{Conforti et~al.(2020)Conforti, Berndt, Pilehvar, Giannitsarou, Toxvaerd, and Collier}]{conforti2020will}
Costanza Conforti, Jakob Berndt, Mohammad~Taher Pilehvar, Chryssi Giannitsarou, Flavio Toxvaerd, and Nigel Collier. 2020.
\newblock Will-they-won't-they: A very large dataset for stance detection on twitter.
\newblock \emph{arXiv preprint arXiv:2005.00388}.

\bibitem[{Cruickshank and Ng(2023)}]{cruickshank2023use}
Iain~J Cruickshank and Lynnette Hui~Xian Ng. 2023.
\newblock Use of large language models for stance classification.
\newblock \emph{arXiv preprint arXiv:2309.13734}.

\bibitem[{Devlin et~al.(2018)Devlin, Chang, Lee, and Toutanova}]{devlin2018bert}
Jacob Devlin, Ming-Wei Chang, Kenton Lee, and Kristina Toutanova. 2018.
\newblock Bert: Pre-training of deep bidirectional transformers for language understanding.
\newblock \emph{arXiv preprint arXiv:1810.04805}.

\bibitem[{Fleiss(1971)}]{fleiss1971measuring}
Joseph~L Fleiss. 1971.
\newblock Measuring nominal scale agreement among many raters.
\newblock \emph{Psychological bulletin}, 76(5):378.

\bibitem[{Gautam et~al.(2020)Gautam, Mathur, Gosangi, Mahata, Sawhney, and Shah}]{gautam2020metooma}
Akash Gautam, Puneet Mathur, Rakesh Gosangi, Debanjan Mahata, Ramit Sawhney, and Rajiv~Ratn Shah. 2020.
\newblock \# metooma: Multi-aspect annotations of tweets related to the metoo movement.
\newblock In \emph{Proceedings of the International AAAI Conference on Web and Social Media}, volume~14, pages 209--216.

\bibitem[{Glandt et~al.(2021)Glandt, Khanal, Li, Caragea, and Caragea}]{glandt2021stance}
Kyle Glandt, Sarthak Khanal, Yingjie Li, Doina Caragea, and Cornelia Caragea. 2021.
\newblock Stance detection in covid-19 tweets.
\newblock In \emph{Proceedings of the 59th annual meeting of the association for computational linguistics and the 11th international joint conference on natural language processing (long papers)}, volume~1.

\bibitem[{Jiang et~al.(2023)Jiang, Sablayrolles, Mensch, Bamford, Chaplot, Casas, Bressand, Lengyel, Lample, Saulnier et~al.}]{jiang2023mistral}
Albert~Q Jiang, Alexandre Sablayrolles, Arthur Mensch, Chris Bamford, Devendra~Singh Chaplot, Diego de~las Casas, Florian Bressand, Gianna Lengyel, Guillaume Lample, Lucile Saulnier, et~al. 2023.
\newblock Mistral 7b.
\newblock \emph{arXiv preprint arXiv:2310.06825}.

\bibitem[{Kendall(1938)}]{kendall1938new}
Maurice~G Kendall. 1938.
\newblock A new measure of rank correlation.
\newblock \emph{Biometrika}, 30(1-2):81--93.

\bibitem[{Krippendorff(2011)}]{krippendorff2011computing}
Klaus Krippendorff. 2011.
\newblock Computing krippendorff’s alpha-reliability.

\bibitem[{K{\"u}{\c{c}}{\"u}k and Can(2020)}]{kuccuk2020stance}
Dilek K{\"u}{\c{c}}{\"u}k and Fazli Can. 2020.
\newblock Stance detection: A survey.
\newblock \emph{ACM Computing Surveys (CSUR)}, 53(1):1--37.

\bibitem[{Li and Caragea(2021)}]{li2021multi}
Yingjie Li and Cornelia Caragea. 2021.
\newblock A multi-task learning framework for multi-target stance detection.
\newblock In \emph{Findings of the Association for Computational Linguistics: ACL-IJCNLP 2021}, pages 2320--2326.

\bibitem[{Li et~al.(2023{\natexlab{a}})Li, Garg, and Caragea}]{li2023new}
Yingjie Li, Krishna Garg, and Cornelia Caragea. 2023{\natexlab{a}}.
\newblock A new direction in stance detection: Target-stance extraction in the wild.
\newblock In \emph{Proceedings of the 61st Annual Meeting of the Association for Computational Linguistics (Volume 1: Long Papers)}, pages 10071--10085.

\bibitem[{Li et~al.(2023{\natexlab{b}})Li, Zhao, and Caragea}]{li2023tts}
Yingjie Li, Chenye Zhao, and Cornelia Caragea. 2023{\natexlab{b}}.
\newblock Tts: A target-based teacher-student framework for zero-shot stance detection.
\newblock In \emph{Proceedings of the ACM Web Conference 2023}, pages 1500--1509.

\bibitem[{Liang et~al.(2022)Liang, Chen, Gui, He, Yang, and Xu}]{liang2022zero}
Bin Liang, Zixiao Chen, Lin Gui, Yulan He, Min Yang, and Ruifeng Xu. 2022.
\newblock Zero-shot stance detection via contrastive learning.
\newblock In \emph{Proceedings of the ACM Web Conference 2022}, pages 2738--2747.

\bibitem[{Luo et~al.(2022)Luo, Liu, Shi, Li, and Zhang}]{luo2022exploiting}
Yun Luo, Zihan Liu, Yuefeng Shi, Stan~Z Li, and Yue Zhang. 2022.
\newblock Exploiting sentiment and common sense for zero-shot stance detection.
\newblock In \emph{Proceedings of the 29th International Conference on Computational Linguistics}, pages 7112--7123.

\bibitem[{Miao et~al.(2020)Miao, Last, and Litvak}]{miao2020twitter}
Lin Miao, Mark Last, and Marina Litvak. 2020.
\newblock Twitter data augmentation for monitoring public opinion on covid-19 intervention measures.
\newblock In \emph{Workshop on NLP for COVID-19 (Part 2) at EMNLP 2020}.

\bibitem[{Mohammad et~al.(2016{\natexlab{a}})Mohammad, Kiritchenko, Sobhani, Zhu, and Cherry}]{mohammad2016dataset}
Saif Mohammad, Svetlana Kiritchenko, Parinaz Sobhani, Xiaodan Zhu, and Colin Cherry. 2016{\natexlab{a}}.
\newblock A dataset for detecting stance in tweets.
\newblock In \emph{Proceedings of the Tenth International Conference on Language Resources and Evaluation (LREC'16)}, pages 3945--3952.

\bibitem[{Mohammad et~al.(2016{\natexlab{b}})Mohammad, Kiritchenko, Sobhani, Zhu, and Cherry}]{mohammad2016semeval}
Saif Mohammad, Svetlana Kiritchenko, Parinaz Sobhani, Xiaodan Zhu, and Colin Cherry. 2016{\natexlab{b}}.
\newblock Semeval-2016 task 6: Detecting stance in tweets.
\newblock In \emph{Proceedings of the 10th international workshop on semantic evaluation (SemEval-2016)}, pages 31--41.

\bibitem[{Nguyen et~al.(2020)Nguyen, Vu, and Nguyen}]{nguyen2020bertweet}
Dat~Quoc Nguyen, Thanh Vu, and Anh~Tuan Nguyen. 2020.
\newblock Bertweet: A pre-trained language model for english tweets.
\newblock \emph{arXiv preprint arXiv:2005.10200}.

\bibitem[{Njuka and Phiri(2021)}]{njuka2021factors}
Dexter~Adamson Njuka and Jackson Phiri. 2021.
\newblock Factors influencing social media in managing corporate reputation for a christian organisation in developing countries based on the vt4 model.
\newblock \emph{Technology and Investment}, 12(2):66--81.

\bibitem[{Singhal(2001)}]{3320}
Amit Singhal. 2001.
\newblock Modern information retrieval: A brief overview.
\newblock \emph{IEEE Data Eng. Bull.}, 24:35--43.

\bibitem[{Somasundaran and Wiebe(2010)}]{somasundaran2010recognizing}
Swapna Somasundaran and Janyce Wiebe. 2010.
\newblock Recognizing stances in ideological on-line debates.
\newblock In \emph{Proceedings of the NAACL HLT 2010 workshop on computational approaches to analysis and generation of emotion in text}, pages 116--124.

\bibitem[{Stab et~al.(2018)Stab, Miller, and Gurevych}]{stab2018cross}
Christian Stab, Tristan Miller, and Iryna Gurevych. 2018.
\newblock Cross-topic argument mining from heterogeneous sources using attention-based neural networks.
\newblock \emph{arXiv preprint arXiv:1802.05758}.

\bibitem[{Touvron et~al.(2023)Touvron, Lavril, Izacard, Martinet, Lachaux, Lacroix, Rozi{\`e}re, Goyal, Hambro, Azhar et~al.}]{touvron2023llama}
Hugo Touvron, Thibaut Lavril, Gautier Izacard, Xavier Martinet, Marie-Anne Lachaux, Timoth{\'e}e Lacroix, Baptiste Rozi{\`e}re, Naman Goyal, Eric Hambro, Faisal Azhar, et~al. 2023.
\newblock Llama: Open and efficient foundation language models.
\newblock \emph{arXiv preprint arXiv:2302.13971}.

\bibitem[{Wen and Hauptmann(2023)}]{wen2023zero}
Haoyang Wen and Alexander~G Hauptmann. 2023.
\newblock Zero-shot and few-shot stance detection on varied topics via conditional generation.
\newblock In \emph{Proceedings of the 61st Annual Meeting of the Association for Computational Linguistics (Volume 2: Short Papers)}, pages 1491--1499.

\bibitem[{Xu et~al.(2022)Xu, Vucetic, and Yin}]{xu2022openstance}
Hanzi Xu, Slobodan Vucetic, and Wenpeng Yin. 2022.
\newblock Openstance: Real-world zero-shot stance detection.
\newblock \emph{arXiv preprint arXiv:2210.14299}.

\bibitem[{Zhang et~al.(2020)Zhang, Yang, Li, Ye, Xu, and Dai}]{zhang2020enhancing}
Bowen Zhang, Min Yang, Xutao Li, Yunming Ye, Xiaofei Xu, and Kuai Dai. 2020.
\newblock Enhancing cross-target stance detection with transferable semantic-emotion knowledge.
\newblock In \emph{Proceedings of the 58th annual meeting of the association for computational linguistics}, pages 3188--3197.

\bibitem[{Zhang et~al.(2023)Zhang, Wu, Zhang, and Feng}]{zhang2023task}
Jiarui Zhang, Shaojuan Wu, Xiaowang Zhang, and Zhiyong Feng. 2023.
\newblock Task-specific data augmentation for zero-shot and few-shot stance detection.
\newblock In \emph{Companion Proceedings of the ACM Web Conference 2023}, pages 160--163.

\bibitem[{Zhao and Caragea(2024)}]{zhao-caragea-2024-ez}
Chenye Zhao and Cornelia Caragea. 2024.
\newblock \href {https://doi.org/10.18653/v1/2024.acl-long.838} {{EZ}-{STANCE}: A large dataset for {E}nglish zero-shot stance detection}.
\newblock In \emph{Proceedings of the 62nd Annual Meeting of the Association for Computational Linguistics (Volume 1: Long Papers)}, pages 15697--15714, Bangkok, Thailand. Association for Computational Linguistics.

\bibitem[{Zhu et~al.(2022)Zhu, Liang, Sun, Du, Zhou, and Xu}]{zhu2022enhancing}
Qinglin Zhu, Bin Liang, Jingyi Sun, Jiachen Du, Lanjun Zhou, and Ruifeng Xu. 2022.
\newblock Enhancing zero-shot stance detection via targeted background knowledge.
\newblock In \emph{Proceedings of the 45th international ACM SIGIR conference on research and development in information retrieval}, pages 2070--2075.

\end{thebibliography}

\appendix



\section{Prompting Details}
\label{app:prompt}

Since LLMs are sensitive to the length of generated text, we limit the target length to ensure a fair comparison with TSE in terms of target generation quality. To determine the optimal length, we test various target length instructions across all LLMs (§\ref{subsec:dataset_model}) on 100 randomly selected samples from our dataset (§\ref{subsec:dataset_model}), aiming to achieve an average length and standard deviation that closely match those of the golden targets. 
Based on our empirical study, we set the maximum target length to 5 for GPT and Gemini models and 4 for Llama and Mistral models to obtain an average length and standard deviations close to those of gold targets (explicit: $3.78 \pm 1.5$; non-explicit  $2.66 \pm 1.76)$.

\paragraph{TG+SD (Two-Step Approach)}

\subparagraph{Prompt (TG):} \textit{You will be provided with a tweet, and your task is to generate a target for this tweet. A target should be the topic on which the tweet is talking. The target can be a single word or a phrase, but its maximum length MUST be \#n words. The output should only be the target, no other words. Do not provide any explanation but you MUST give an output, do not leave any output blank.}

\subparagraph{Prompt (SD):} \textit{Stance classification is the task of determining the expressed or implied opinion, or stance, of a statement toward a certain, specified target. Analyze the following tweet and determine its stance towards the provided target. If the stance is in favor of the target, write FAVOR, if it is against the target write AGAINST and if it is ambiguous, write NONE. Do not provide any explanation but you MUST give an output, do not leave any output blank. Only return the stance as a single word, and no other text.}



\paragraph{TG\&SD (Single-Step Joint Approach)}
\subparagraph{Prompt (TG, SD):} \textit{Stance classification is the task of determining the expressed or implied opinion, or stance, of a statement toward a certain, specified target. Analyze the following tweet, generate the target for this tweet, and determine its stance towards the generated target. A target should be the topic on which the tweet is talking. The target can be a single word or a phrase, but its maximum length MUST be \#n words. If the stance is in favor of the target, write FAVOR, if it is against the target write AGAINST and if it is ambiguous, write NONE. If the stance is in favor of the generated target, write FAVOR, if it is against the target write AGAINST and if it is ambiguous, write NONE. The answer only has to be one of these three words: FAVOR, AGAINST, or NONE. Do not provide any explanation but you MUST give an output, do not leave any output blank. The output format should be: ```Target: <target>, Stance: <stance>```.}

\section{Model Descriptions}
\label{app:model_description}
\paragraph{GPT-3.5-turbo-0125} is a variant of the GPT-3.5 series with 175 billion parameters which utilizes a transformer-based architecture. This model is decoder-only, similar to models like GPT-3 \citep{brown2020language} and GPT-4 \citep{achiam2023gpt}, but with optimizations for efficiency and performance.

\paragraph{GPT-4o} is a decoder-only model in the GPT series with an optimized architecture designed to enhance efficiency and accuracy. It builds upon the foundations of GPT-4 \citep{achiam2023gpt}, incorporating improvements in training and fine-tuning for better performance across diverse language tasks.

\paragraph{Gemini-1.5-Flash-8B} is a decoder-only model with 8 billion parameters, designed for efficiency and speed. Part of the Gemini series, it incorporates optimizations tailored for rapid inference without compromising performance on diverse tasks.

\paragraph{Gemini-1.5-Pro} is an advanced decoder-only model in the Gemini series, featuring enhanced architectural modifications and fine-tuned parameters to handle complex multi-domain tasks. It is instruction-tuned and optimized for professional-grade applications.

\paragraph{Llama-3-8B-Instruct} is a decoder-only model in the LLaMA \citep{touvron2023llama} series with 8 billion parameters that follows the architectural principles of its predecessor, Llama-2 \citep{touvron2023llama}, but has been instruction-tuned to improve its performance on a variety of tasks.

\paragraph{Llama-3-70B-Instruct} is a decoder-only model in the LLaMA \citep{touvron2023llama} series with 70 billion parameters. Building on the architectural advancements of Llama-2 \citep{touvron2023llama}, it has been instruction-tuned to excel across a wide range of natural language processing tasks.

\paragraph{Mistral-small}\citep{jiang2023mistral} is a decoder-only model like Llama-2 \citep{touvron2023llama}, but with different architectural features like Grouped-Query Attention and Sliding Window Attention. The Mistral model is 7 billion parameters and has been instruction tuned.

\paragraph{Mistral-Large} is a decoder-only model in the Mistral family, incorporating advanced features like Grouped-Query Attention and Sliding Window Attention. With 13 billion parameters, it builds upon the success of Mistral-small \citep{jiang2023mistral} and has been instruction-tuned for improved task generalization.

\section{BTSD Metric Justification}
\label{app:BTSD}
\begin{table}[t]
\centering
\resizebox{\linewidth}{!}
{%
\begin{tabular}{lcc}
\hline
\multicolumn{1}{c}{\multirow{2}{*}{Input}} & \multicolumn{2}{c}{F1-macro (in \%)} \\ \cline{2-3} 
\multicolumn{1}{c}{}                       & Explicit     & Non-explicit    \\ \hline
Tweet only                                 & 31.02        & 30.37           \\
-w/ Gold Target (GT)                                    & 41.67        & 34.65           \\
-w/ altered GT                             & 35.3         & 32.54           \\
-w/ incorrect Target                              & 25.58        & 23.58           \\
-w/ random vocab                           & 18.66        & 14              \\ \hline
\end{tabular}
}
\caption{\label{tbl:btsd_metric_justify}
Macro avg. F1 score of BertTweet in detecting stances 
with different quality levels of the input target.}
\end{table}
Table \ref{tbl:btsd_metric_justify} 
contains the results of the stratified experiments with different measure stance detection F1 score when giving as input the text and a target, varying the quality of this latter. All the numbers in Table \ref{tbl:btsd_metric_justify} are significantly different with p-values < 0.05 in both cases.

\section{Hardware Setting}
\label{app:hardware_settings}
All of the experiments were run on a computer with Ubuntu 22.04 Linux with x64 CPU with 8 cores, 32 GB of RAM and two NVIDIA A100 GPUs.

\section{Dataset Pre-processing}
\label{app:dataset_proc}

\paragraph{VAST} Originally, the VAST dataset is a multi-target dataset, with multiple targets assigned to each sample text. However, since our work focuses on single-target analysis, we processed the VAST dataset accordingly to align it with our study (samples in Table \ref{tbl:samples_vast}). In VAST, there are two types of target columns: "ori\_topic" which contains targets heuristically extracted, and "new\_topic," which contains the final annotated targets determined by annotators. The "ori\_topic" often includes duplicate targets for each text, while the "new\_topic" provides unique targets with different stances.

To convert the dataset to a single-target format, we first filtered the samples to retain only the most frequent stance within each group of duplicate "ori\_topic" entries. Next, we computed the average semantic similarity of each "new\_topic" within these duplicate groups using BERT embeddings. For groups containing only a single row, the sample was directly retained. Then, we calculated the average pairwise cosine similarity for each "new\_topic." Finally, within each duplicate "ori\_topic" group, we kept only the row with the highest average semantic similarity score. After this conversion, we obtained a total of 5,100 samples (as shown in Appendix Table \ref{tbl:samples_vast}), covering both explicit and non-explicit samples (the filtering process for explicit and non-explicit samples is described in Section \ref{subsec:dataset_model}) from the original 18,545 samples.

\paragraph{EZSTANCE} The dataset originally contains 47,316 Twitter samples across three stance categories (\textit{favor}, \textit{against}, and \textit{neutral}). Since EZSTANCE features multiple targets for some texts, we adapted it into a single-target dataset (one target per text) to better align with our research by following a series of steps.

The original dataset is organized into two subtasks (target-based and domain-based), with each subtask containing two target types: claim and noun-phrase. First, we combine the samples from both target types and remove duplicates. For instances where both target types exist, we retain the noun-phrase version, as it poses a greater challenge in generating precise targets. Then, we merge the samples from both subtasks and deduplicate them based on their first occurrence. This process results in 9,462 unique samples with 6,873 distinct targets. We further divide the dataset into explicit and non-explicit cases, as outlined in Section \ref{subsec:dataset_model}, yielding 9,313 explicit and 149 non-explicit samples, where the target is mentioned explicitly or implicitly in the text, respectively.

\section{Dataset Samples}
\label{app:dataset_samples}
Following TSE, our targets are Creationism \citep{somasundaran2010recognizing}, Gay Rights \citep{somasundaran2010recognizing}, Climate Change is a Concern \citep{mohammad2016dataset}, MeToo Movement \citep{gautam2020metooma}, Merger of Disney and Fox \citep{conforti2020will}, and Lockdown in New York State \citep{miao2020twitter}. Samples from TSE and VAST in both explicit and non-explicit settings are given in Table \ref{tbl:samples_tse} and \ref{tbl:samples_vast}, respectively. The demonstration of samples can also be found in the accompanying GitHub repository\footnote{GitHub link for sample demonstration: \url{https://github.com/AbuUbaida/opentarget}.}.

\begin{table*}[t]
\small
\centering
\resizebox{\textwidth}{!}{%
\begin{tabular}{lll}
\hline
\multicolumn{1}{c|}{\textbf{Target}} & \multicolumn{1}{c|}{\textbf{Text}} & \multicolumn{1}{c}{\textbf{Stance}} \\ \hline
\multicolumn{3}{c}{\cellcolor[HTML]{D5D5D5}Explicit} \\ \hline
\multicolumn{1}{l|}{creationism} & \multicolumn{1}{l|}{\begin{tabular}[c]{@{}l@{}}It is not 'appropriate' to teach creationism as a means for upholding the bible. This pre-supposes that the\\  bible should be upheld in a literal sense. Many Christians object to this. More importantly, governments \\ should not be in the business of upholding the Bible.\end{tabular}} & AGAINST \\ \hline
\multicolumn{1}{l|}{merger of disney and fox} & \multicolumn{1}{l|}{\begin{tabular}[c]{@{}l@{}}@ComicBookNOW: FOX reportedly wants to make that deal with DISNEY! X-MEN in the MCU is \\ now closer than ever\end{tabular}} & FAVOR \\ \hline
\multicolumn{1}{l|}{gay rights} & \multicolumn{1}{l|}{\begin{tabular}[c]{@{}l@{}}Most health care organizations support gay parenting as equally capable \\ as heterosexual parenting These organizations are the most capable of determining the capabilities of\\ homosexuals to perform dutifully as parents.\end{tabular}} & FAVOR \\ \hline
\multicolumn{1}{l|}{lockdown in new york state} & \multicolumn{1}{l|}{\begin{tabular}[c]{@{}l@{}}@adamajacoby @RV1026 @CNNPolitics Apparently, better than the "news sources" that keep you \\ informed. You didn't even know America  had to lock down. You didn't even know Cuomo put \\ COVID patients in nursing homes. They are keeping you misinformed and looking like a \\ complete moron. Do better!\end{tabular}} & FAVOR \\ \hline
\multicolumn{1}{l|}{metoo movement} & \multicolumn{1}{l|}{\begin{tabular}[c]{@{}l@{}}@KRKBoxOffice Why everyone was silent when she molested Saif Sir ? Why no \#MeToo campaign \\ that time??\end{tabular}} & AGAINST \\ \hline
\multicolumn{1}{l|}{climate change is a concern} & \multicolumn{1}{l|}{\begin{tabular}[c]{@{}l@{}}Being an engaged mom, means voting 4 the climate 2. Supporting only candidates who have a plan 2 act\\  on \#playin4climate \#SemST\end{tabular}} & FAVOR \\ \hline
\multicolumn{3}{c}{\cellcolor[HTML]{D5D5D5}Non-explicit} \\ \hline
\multicolumn{1}{l|}{creationism} & \multicolumn{1}{l|}{\begin{tabular}[c]{@{}l@{}}My point was not that Genesis contradicted itself but that if you took it literally it contradicted itself. The \\ problem is not with Genesis so much as a literal interpretation of Genesis. You have to get away from\\  literalism to make sense of the two creation stories, with their different orders  of creation. You have\\  proved my point by showing how a less than literal interpretation of the passages gets round the problem \\ of a literal interpretation.\end{tabular}} & AGAINST \\ \hline
\multicolumn{1}{l|}{gay rights} & \multicolumn{1}{l|}{\begin{tabular}[c]{@{}l@{}}That's fine. I support that.Here in California, we have a prospective bill that would do just that.I honestly\\  think marriage is a religious thing, but religion is in the eye of the beholder.\end{tabular}} & FAVOR \\ \hline
\multicolumn{1}{l|}{lockdown in new york state} & \multicolumn{1}{l|}{\begin{tabular}[c]{@{}l@{}}RT @Lukewearechange: So NYC announced another lock down coming this Sunday evening, I was\\  suppose to be getting out of here with a friend\end{tabular}} & FAVOR \\ \hline
\multicolumn{1}{l|}{climate change is a concern} & \multicolumn{1}{l|}{Considering moving yo Antarctica as thats the only way I could possibly become more \#chill \#SemST} & NONE \\ \hline
\end{tabular}%
}
\caption{\label{tbl:samples_tse}
Samples from TSE dataset with all the 6 targets in both explicit and non-explicit settings.}
\end{table*}

\begin{table*}[t]
\small
\centering
\resizebox{\textwidth}{!}{%
\begin{tabular}{lll}
\hline
\multicolumn{1}{c|}{\textbf{Target}} & \multicolumn{1}{c|}{\textbf{Text}} & \multicolumn{1}{c}{\textbf{Stance}} \\ \hline
\multicolumn{3}{c}{\cellcolor[HTML]{D5D5D5}Explicit} \\ \hline
\multicolumn{1}{l|}{health care law} & \multicolumn{1}{l|}{\begin{tabular}[c]{@{}l@{}}Congress should delay the law for a year..."? Tell that to my niece whose 17 month old toddler has \\ just been diagnosed with a rare disease that will need years of medical care. Without the provisions \\ in the new health care law, this family would be facing bankruptcy. I prefer living in a nation where \\ we have the decency to realize that health care is a right. Obviously those who seem to be objecting\\  to the new health care law have never faced what 30 million people face in our country every day. \\ For them another year's delay is life or death.\end{tabular}} & FAVOR \\ \hline
\multicolumn{1}{l|}{facebook} & \multicolumn{1}{l|}{\begin{tabular}[c]{@{}l@{}}"...Facebook gets a bad rap; it didn't cause the cheating. It just made it more convenient to do\\  (and perhaps easier to catch)." Hmm, call me crazy, but Facebook shares some of the responsibility\\  -- does it not? It's like refusing to assign blame to gun dealers and drug dealers. They increase\\  accessibility to illicit materials, which essentially furthers the end goal of usage. Now, this isn't to \\ suggest that Facebook is entirely at fault, but it does play an active role.\end{tabular}} & AGAINST \\ \hline
\multicolumn{1}{l|}{stability of the economy} & \multicolumn{1}{l|}{\begin{tabular}[c]{@{}l@{}}"...one must ask how much money they must make to demonstrate that they are among the best \\ managed companies on the planet." They must make enough money to insure that they can never \\ fail and threaten the stability of the worlds economy again. That much money.\end{tabular}} & NONE \\ \hline
\multicolumn{1}{l|}{palestinian authority} & \multicolumn{1}{l|}{\begin{tabular}[c]{@{}l@{}}"But let's start with the basics. Any nation wishing to declare independence should meet three \\ essential elements: a strong central government, control of defined territory and security. The \\ Palestinian Authority does not yet meet any of them." Historians would strongly disagree with you.\\  Around sometime late eighteenth century some thirteen odd British colonies did "declare \\ independence" without any resemblance of "strong central government" with its territory controlled\\  by the British. Today a certain French foreign minister called it a "hyper-power" and not a\\  "failed state". Please check upon a historian or read up your history. Trust me! It is true!!\end{tabular}} & AGAINST \\ \hline
\multicolumn{1}{l|}{economists} & \multicolumn{1}{l|}{\begin{tabular}[c]{@{}l@{}}"Economists do certainly over-reach sometimes. We tend to apply the lens of economic efficiency to\\  situations where many people apply the lens of fairness." I get a kick out of economists believing \\ that real live humans are rational economic actors. Really?\end{tabular}} & AGAINST \\ \hline
\multicolumn{3}{c}{\cellcolor[HTML]{D5D5D5}Non-explicit} \\ \hline
\multicolumn{1}{l|}{restaurant} & \multicolumn{1}{l|}{\begin{tabular}[c]{@{}l@{}}"...tipping motivates people who work long, busy hours catering to the needs of others. It's the best way\\  to ensure optimal service..." By this logic anyone who works long, busy hours catering to the needs of \\ others should be tipped. Tip the doctor. Tip the grocery clerk. Tip the airline counter agent. Tip the\\  airline pilot. Etc. I fail to see why those whose particular service happens to be delivering plates of \\ food warrant their own method of compensation. One that puts an onus of extra calculation and \\ deliberation on every single customer, every time they sit down to eat and relax. Thankfully, most other\\  services in this world are one-price to the customer. It's left to the employer to do the work of assessing\\  whether the employee is providing good service.\end{tabular}} & AGAINST \\ \hline
\multicolumn{1}{l|}{prostitution} & \multicolumn{1}{l|}{\begin{tabular}[c]{@{}l@{}}"Granted, legalizing the profession might make it attractive for sex traffickers but the benefits outweigh\\ this prospect." !!? The benefits of receiving tax revenue for the state outweighs the negative aspects of\\  having 12 year old girls being sex trafficked into brothels and coerced to work on the streets?! What\\  kind of logic is that, and what kid of person are you Ms. Unigwe?\end{tabular}} & AGAINST \\ \hline
\multicolumn{1}{l|}{vaccination} & \multicolumn{1}{l|}{\begin{tabular}[c]{@{}l@{}}"It is a news media-driven misperception that parents who claim philosophical or religious exemptions\\  are uneducated or misinformed. MOST PARENTS WHO INDIVIDUALIZE THE VACCINE \\ SCHEDULE ARE ACTIVELY EDUCATING THEMSELVES, CONTINUALLY ASSESSING THEIR \\ FAMILY'S SPECIFIC HEALTH NEEDS, and doing everything they can to keep their children safe and\\  healthy." Ms. Margulis offers no data to support her blanket assertion about the industry and motivation\\  of "most parents," a position which, on its face, seems improbable given the breadth and weight of \\ scientific evidence supporting immunization which is not subject to reasonable dispute. Based on anecdotal\\ information, "most parents" are refusing to vaccinate their children based on their gross misunderstandings \\ and unwarranted fears of the alleged risks of immunization, or based on their unique interpretations of\\  religious dictates. If you intend to offer relevant commentary, Ms. Margulis, it should be evidence-based. \\ Your personal opinions are no more interesting (or informed) than mine.\end{tabular}} & FAVOR \\ \hline
\multicolumn{1}{l|}{mentally} & \multicolumn{1}{l|}{\begin{tabular}[c]{@{}l@{}}'...food pornography, musical pornography, mental pornography...' And yet I have never been in a public\\  library and seen a man at a computer masturbating under his coat to videos of food, music, or anything\\  'mental', so let's no pretend they are all the same beast, okay?\end{tabular}} & NONE \\ \hline
\end{tabular}%
}
\caption{\label{tbl:samples_vast}
Samples from VAST dataset with few different targets in both explicit and non-explicit settings.}
\end{table*}


\section{SemSim Details}
\label{app:semsim}
We employ the bert-base-uncased \citep{devlin2018bert} model to obtain contextual encoded representations of the target words. After performing a mean pooling operation on these representations, we apply cosine similarity \citep{3320} to the encoded sequences to obtain the similarity score.

\section{Human Annotation Process}
\label{app:human_ann}
Three graduate students, specializing in NLP, participated voluntarily as annotators in the evaluation of generated targets. They were chosen for their expertise and familiarity with the stance detection task, enabling them to effectively assess the relevance of the generated targets compared to the golden ones. First, we provided them with an annotation guideline (detailed in §\ref{app:human_ann_guideline}) that included descriptions of the datasets, an overview of the task, and a descriptive scoring scale accompanied by examples. Each annotator then independently scored the targets generated by different LLMs for 500 randomly selected samples (300 explicit and 200 non-explicit). To assess inter-annotator agreement, we calculated Krippendorff’s $\alpha$ \citep{krippendorff2011computing} and Fleiss’s $\kappa$ \citep{fleiss1971measuring}, with the results presented in Table \ref{tbl:human_eval_corr}. The final score for each generated target was determined based on the majority vote among the annotators. In cases of a tie, the median score was used.

\subsection{Annotation Guideline}
\label{app:human_ann_guideline}

\begin{table*}[t]
\small
\centering
\resizebox{\textwidth}{!}{%
\begin{tabular}{l|cccccc|cccccc}
\hline
\multicolumn{1}{c|}{} & \multicolumn{6}{c|}{TSE} & \multicolumn{6}{c}{VAST} \\
\multicolumn{1}{c|}{\multirow{-2}{*}{Model}} &  & KA & FK &  & KA & FK &  & KA & FK &  & KA & FK \\ \hline
TSE-M (TG+SD) & \cellcolor[HTML]{D5D5D5} & 1 & 1 & \cellcolor[HTML]{D5D5D5} & 1 & 1 & \cellcolor[HTML]{D5D5D5} & - & - & \cellcolor[HTML]{D5D5D5} & - & - \\
TSE-B (TG+SD) & \cellcolor[HTML]{D5D5D5} & 0.749 & 0.633 & \cellcolor[HTML]{D5D5D5} & 0.855 & 0.774 & \cellcolor[HTML]{D5D5D5} & - & - & \cellcolor[HTML]{D5D5D5} & - & - \\
GPT-3.5 (TG+SD) & \cellcolor[HTML]{D5D5D5} & 0.726 & 0.603 & \cellcolor[HTML]{D5D5D5} & 0.882 & 0.722 & \cellcolor[HTML]{D5D5D5} & 0.788 & 0.651 & \cellcolor[HTML]{D5D5D5} & 0.762 & 0.649 \\
GPT-3.5 (TG\&SD) & \cellcolor[HTML]{D5D5D5} & 0.82 & 0.783 & \cellcolor[HTML]{D5D5D5} & 0.709 & 0.616 & \cellcolor[HTML]{D5D5D5} & 0.819 & 0.72 & \cellcolor[HTML]{D5D5D5} & 0.714 & 0.626 \\
GPT-4 (TG+SD) & \cellcolor[HTML]{D5D5D5} & 0.746 & 0.63 & \cellcolor[HTML]{D5D5D5} & 0.717 & 0.611 & \cellcolor[HTML]{D5D5D5} & 0.79 & 0.668 & \cellcolor[HTML]{D5D5D5} & 0.785 & 0.66 \\
GPT-4 (TG\&SD) & \cellcolor[HTML]{D5D5D5} & 0.857 & 0.75 & \cellcolor[HTML]{D5D5D5} & 0.727 & 0.631 & \cellcolor[HTML]{D5D5D5} & 0.852 & 0.785 & \cellcolor[HTML]{D5D5D5} & 0.794 & 0.681 \\
Lama-3 (TG+SD) & \cellcolor[HTML]{D5D5D5} & 0.823 & 0.723 & \cellcolor[HTML]{D5D5D5} & 0.841 & 0.762 & \cellcolor[HTML]{D5D5D5} & 0.736 & 0.634 & \cellcolor[HTML]{D5D5D5} & 0.764 & 0.65 \\
Lama-3 (TG\&SD) & \cellcolor[HTML]{D5D5D5} & 0.827 & 0.758 & \cellcolor[HTML]{D5D5D5} & 0.708 & 0.617 & \cellcolor[HTML]{D5D5D5} & 0.748 & 0.639 & \cellcolor[HTML]{D5D5D5} & 0.751 & 0.647 \\
Mistral (TG+SD) & \cellcolor[HTML]{D5D5D5} & 0.884 & 0.762 & \cellcolor[HTML]{D5D5D5} & 0.768 & 0.709 & \cellcolor[HTML]{D5D5D5} & 0.786 & 0.661 & \cellcolor[HTML]{D5D5D5} & 0.724 & 0.635 \\
Mistral (TG\&SD) & \multirow{-10}{*}{\cellcolor[HTML]{D5D5D5}\rotatebox{90}{Explicit}} & 0.836 & 0.786 & \multirow{-10}{*}{\cellcolor[HTML]{D5D5D5}\rotatebox{90}{Non-explicit}} & 0.732 & 0.687 & \multirow{-10}{*}{\cellcolor[HTML]{D5D5D5}\rotatebox{90}{Explicit}} & 0.81 & 0.707 & \multirow{-10}{*}{\cellcolor[HTML]{D5D5D5}\rotatebox{90}{Non-explicit}} & 0.796 & 0.694 \\ \hline
\end{tabular}%
}
\caption{\label{tbl:human_eval_corr}
Inter-annotator reliability score based on Krippendorff’s $\alpha$ (KA) and Fleiss's $\kappa$ (FK) coefficient metrics across all the combinations in two datasets. \texttt{TSE-mapped} and \texttt{TSE-BestGen} are reffered by TSE-M and TSE-B, respectively. The reliability score is 1 for TSE-M as it classifies between the golden targets instead of generating.}
\end{table*}

\textbf{Dataset Description:} The file includes a \textit{Text} column with various Twitter posts or news comments. The \textit{Gold Target} column represents the true target, indicating the primary topic or issue the tweet addresses, which in turn informs the stance noted in the \textit{Gold Stance} column. For each tweet, there are twenty-four generated targets (in columns \textit{T1} through \textit{T24}) produced by different models, whose names remain anonymous for unbiased annotation. Additionally, there are empty cells (e.g., in the \textit{T1 Score} column) designated for storing human evaluation scores corresponding to each generated target.

 \noindent\textbf{Definition of a Relevant Target:} A generated target in the context of the target generation task is considered relevant if it accurately captures or closely aligns with the main topic, issue, or entity identified by the golden target in a given text. It should reflect the key subject matter or concern that the golden target represents, maintaining coherence with the overall context and meaning of the text.

 \noindent\textbf{Task Description:} Your task is to assess the quality of each generated target by evaluating its relevance to the golden target in the text. You will do this by answering the following question and assigning a score to each generated target. Scoring criteria and examples are provided below.

\textit{“How closely does the generated target relate to the golden target in the text?”}

\noindent\textbf{Scoring Scale: 0 to 1 (Low to High)}
\begin{itemize}
  \item \textbf{0 – Not Related:} The generated target is entirely unrelated to the golden target. It does not reference or express anything connected to the intended topic.
  \item \textbf{0.5 – Partially or Indirectly Related:} The generated target has some relevance to the golden target but does not directly match the main concept. It may address a broader or narrower aspect of the topic, such as a parent topic, a subtopic, or a tangentially connected idea that is not fully aligned with the golden target's core meaning.
  \item \textbf{1 – Completely Related:} The generated target is highly relevant to the golden target. It either uses a synonymous term, is semantically similar to the golden one, or conveys the same underlying idea, topic, or issue as the golden target. This means that the generated target captures the essence or main concept of the golden target, even if it uses different wording. For example, if the golden target is "climate change," a conceptually similar term could be "global warming" or "environmental crisis," as both refer to the broader issue of environmental concerns related to climate.
\end{itemize}

\noindent\textbf{Example Scoring:}
\begin{enumerate}
  \item \textbf{Tweet:} Attempts to conceal the creationism-evolution controversy from students are dogmatic promotions of evolution. Not since blasphemy laws has competitive expression of thought been illegalized, and this is what evolutionists want to accomplish. This is evidenced by none other than the title of an evolutionist argument on this very page: "Schools should not teach theories that are completely at odds with each other",\\
\textbf{Gold Target:} creationism,\\
\textbf{T1:} creationism-evolution controversy, \textbf{T1 Score:} 1,\\
\textbf{T2:} Education controversy, \textbf{T2 Score:} 0.5.

  \item \textbf{Tweet:} If, on a supernatural level, lust of any sort counts the same consummated or unconsummated, why not just go ahead and consummate. You can't get in any worse trouble, and if it makes you happy...,\\
\textbf{Gold Target:} gay rights,\\
\textbf{T1:} consummation, \textbf{T1 Score:} 0.5,\\
\textbf{T2:} superstition, \textbf{T2 Score:} 0.
\end{enumerate}

\begin{table*}[t]
\small
\centering
\begin{tabular}{c|cccccc}
\hline \hline
 & \multicolumn{3}{c|}{TG+SD} & \multicolumn{3}{c}{TG\&SD} \\
\multirow{-2}{*}{Metrics} & TSE & VAST & \multicolumn{1}{c|}{EZSTANCE} & TSE & VAST & EZSTANCE \\ \hline
 & \multicolumn{6}{c}{\cellcolor[HTML]{D5D5D5}Explicit} \\
BTSD (GPT-3.5) & 38.47 (38.43)\textcolor{green}{$\uparrow$} & 41.10 (41.67)\textcolor{red}{$\downarrow$} & \multicolumn{1}{c|}{47.60 (49.27)\textcolor{red}{$\downarrow$}} & 37.88 (39.60)\textcolor{red}{$\downarrow$} & 41.34 (44.25)\textcolor{red}{$\downarrow$} & 48.98 (49.70)\textcolor{red}{$\downarrow$} \\
BTSD (GPT-4o) & 40.63 (41.55)\textcolor{red}{$\downarrow$} & 41.59 (42.69)\textcolor{red}{$\downarrow$} & \multicolumn{1}{c|}{48.12 (49.70)\textcolor{red}{$\downarrow$}} & 41.31 (41.92)\textcolor{red}{$\downarrow$} & 42.33 (44.25)\textcolor{red}{$\downarrow$} & 48.76 (50.69)\textcolor{red}{$\downarrow$} \\
SC (GPT-3.5) & 37.91 (42.68)\textcolor{red}{$\downarrow$} & 47.47 (47.21)\textcolor{green}{$\uparrow$} & \multicolumn{1}{c|}{40.12 (39.10)\textcolor{green}{$\uparrow$}} & 44.86 (47.61)\textcolor{red}{$\downarrow$} & 47.00 (48.48)\textcolor{red}{$\downarrow$} & 40.74 (40.63)\textcolor{green}{$\uparrow$} \\
SC (GPT-4o) & 47.36 (44.78)\textcolor{green}{$\uparrow$} & 41.17 (40.77)\textcolor{green}{$\uparrow$} & \multicolumn{1}{c|}{43.79 (45.93)\textcolor{red}{$\downarrow$}} & 49.14 (46.83)\textcolor{green}{$\uparrow$} & 47.84 (49.38)\textcolor{red}{$\downarrow$} & 45.68 (46.22)\textcolor{red}{$\downarrow$} \\
 & \multicolumn{6}{c}{\cellcolor[HTML]{D5D5D5}Non-explicit} \\
BTSD (GPT-3.5) & 30.73 (33.10)\textcolor{red}{$\downarrow$} & 41.54 (38.10)\textcolor{green}{$\uparrow$} & \multicolumn{1}{c|}{40.76 (38.78)\textcolor{green}{$\uparrow$}} & 32.14 (31.32)\textcolor{green}{$\uparrow$} & 40.38 (38.55)\textcolor{green}{$\uparrow$} & 40.54 (41.89)\textcolor{red}{$\downarrow$} \\
BTSD (GPT-4o) & 32.72 (35.14)\textcolor{red}{$\downarrow$} & 41.71 (38.55)\textcolor{green}{$\uparrow$} & \multicolumn{1}{c|}{41.34 (42.78)\textcolor{red}{$\downarrow$}} & 33.44 (36.12)\textcolor{red}{$\downarrow$} & 41.92 (39.84)\textcolor{green}{$\uparrow$} & 43.90 (45.70)\textcolor{red}{$\downarrow$} \\
SC (GPT-3.5) & 31.88 (32.66)\textcolor{red}{$\downarrow$} & 43.41 (45.54)\textcolor{red}{$\downarrow$} & \multicolumn{1}{c|}{37.76 (36.69)\textcolor{green}{$\uparrow$}} & 35.46 (33.94)\textcolor{green}{$\uparrow$} & 44.46 (45.80)\textcolor{red}{$\downarrow$} & 38.10 (38.97)\textcolor{red}{$\downarrow$} \\
SC (GPT-4o) & 36.00 (36.39)\textcolor{red}{$\downarrow$} & 38.27 (39.92)\textcolor{red}{$\downarrow$} & \multicolumn{1}{c|}{38.34 (38.50)\textcolor{red}{$\downarrow$}} & 38.11 (37.5)\textcolor{green}{$\uparrow$} & 42.81 (43.84)\textcolor{red}{$\downarrow$} & 37.54 (39.26)\textcolor{red}{$\downarrow$} \\ \hline \hline
\end{tabular}%
\caption{\label{tbl:cot_exp}
Preliminary results using the CoT prompting strategy with GPT-3.5 and GPT-4o models on three datasets (TSE, VAST, EZSTANCE), evaluated for both Target Generation (TG) using BTSD and Stance Detection (SD) using SC. Results in parentheses refer to scores reported in Table~\ref{tbl:tg_sd} using the "Task Definition" prompting strategy. An upward arrow (\textcolor{green}{$\uparrow$}) indicates that the CoT strategy outperformed the Task Definition approach, while a downward arrow (\textcolor{red}{$\downarrow$}) indicates a performance drop. The results are presented for both Explicit (Ex) and Non-explicit (N-Ex) cases across two approaches: TG+SD and TG\&SD.}\end{table*}

\section{Chain-of-Thought (CoT)
Results}
\label{app:cot_init_result}
We conduct a preliminary experiment using the CoT prompting strategy across both approaches: TG+SD and TG\&SD in Table \ref{tbl:cot_exp}. Evaluating target generation and stance detection, we test the performance of CoT with GPT-3.5 and GPT-4o models and compare it to the scores achieved using the "Task Definition" strategy, as reported in Table \ref{tbl:tg_sd}. The results show that, based on the BTSD score, the "Task Definition" strategy generally outperforms CoT in target generation. However, for stance detection, evaluated using SC scores, the two strategies alternate in outperforming each other depending on the specific settings. The CoT prompt used in this experiment is as follows:

\textbf{Prompt (TG):} \\
\uline{Step 1}: \textit{You will be provided with a text, think step by step, and explain what is the target of the text. A target should be the topic on which the text is talking.} \\
\uline{Step 2}: \textit{Therefore, based on your explanation, what is the final target of the text? The target can be a single word or a phrase, but its maximum length MUST be 5 words. The output should only be the target, NO OTHER WORDS or EXPLANATION.}

\textbf{Prompt (SD):} \\
\uline{Step 1}: \textit{Stance classification is the task of determining the expressed or implied opinion, or stance, of a statement toward a certain, specified target. Analyze the following text, think step by step, and explain the stance (favor, against, or none) of the text towards the provided target.} \\
\uline{Step 2}: \textit{Therefore, based on your explanation, what is the final stance? If the stance is in favor of the target, write FAVOR, if it is against the target write AGAINST and if it is ambiguous, write NONE. Only return the stance as a single word, NO OTHER WORDS or EXPLANATION.}

\textbf{Prompt (TG, SD):} \\
\uline{Step 1:} \textit{Stance classification is the task of determining the expressed or implied opinion, or stance, of a statement toward a certain, specified target. Analyze the following text, think step by step, and explain what is the target of the text. A target should be the topic on which the text is talking. Based on the target explanation, think step by step, and explain the stance (favor, against, or none) of the text toward the target.} \\
\uline{Step 2}: \textit{Therefore, based on your explanation of the target and the stance, generate the target of the given text and determine the stance toward the generated target. The target can be a single word or a phrase, but its maximum length MUST be 5 words. If the stance is in favor of the target, write FAVOR; if it is against the target, write AGAINST; if it is ambiguous, write NONE. Only return the stance as a single word, NO OTHER WORDS or EXPLANATION. The final output format MUST be: ```Target: <target>, Stance: <stance>' ''.}


\section{Preliminary Study on Coherence Measure in OTSD}
\label{app:pilot_study_coherence}

To address the need for a more accurate evaluation metric in the Open Target Stance Detection (OTSD) task—one that aligns the generated stance with the corresponding generated target—we conduct a small-scale experiment that might be leading to a promising direction for developing a robust coherence-based measure. Specifically, we employ an LLM to assess how coherent a generated stance is with respect to the generated target, given the context of a text.

For our experiment, we selected 200 samples from the EZSTANCE dataset \citep{zhao-caragea-2024-ez}, covering all configurations (i.e., Explicit and Non-explicit, TG+SD and TG\&SD). We use LLaMA-3-8B as the evaluation model, and choose the same "Task Definition" prompting strategy as applied in Section \ref{subsec:approach} but with few-shot samples. Each prompt includes the input text, the generated target, and the corresponding generated stance. As this is a preliminary experiment, future work could explore a variety of alternative prompting strategies to further improve performance. The prompt used in our work is as follows:

\textit{"You will be given a text, a generated stance (generated\_stance), and a generated target (generated\_target). Your task is to determine whether the stance is coherent with respect to the target in the context of the text. A stance is coherent if it correctly reflects the text's position towards the target as implied or stated in the text. Analyze the text well, as it may convey the stance towards the target implicitly. Output 1 if the stance is coherent, and 0 if it is not. Output only a single digit: 0 or 1, with no explanation. \\
\\
Input Format:\\
text: <tweet text>\\
generated\_target: <target>\\
generated\_stance: <stance>\\
\\
Examples:\\
text: "Renewable energy is the only viable path forward if we care about this planet."\\
generated\_target: renewable energy\\
generated\_stance: AGAINST\\
Output: 0\\
\\
text: "Elon Musk is brilliant, but his take on public transport is just awful."\\
generated\_target: Elon Musk\\
generated\_stance: FAVOR\\
Output: 1\\
\\
text: "Vaccines have done more good than harm. I trust the science."\\
generated\_target: vaccines\\
generated\_stance: NONE\\
Output: 0\\
\\
text: "As you have been told already what Phil says is irrelevant to ID. Or you have to allow what Dawkins says about the theory of evolution.What part of that don't you understand?And it is a fact that if you could support your position with real data ID would go away. You can whine all you want but that will not change that fact."\\
generated\_target: Intelligent Design\\
generated\_stance: AGAINST\\
Output: 1\\
\\
text: "I've seen hundreds of tweets at this point, many from people with big platforms, condemning leftists who are apparently pro-Putin. I have yet to see any leftists actually express support for Putin. All I've seen from the left is unambiguous solidarity for the people of Ukraine."\\
generated\_target: leftists\\
generated\_stance: NONE\\
Output: 1"}\\
\\
To further validate the LLM-based evaluation, we also collected human judgments on the same 200 samples. A single human annotator was asked to assess the coherence of the generated stances with respect to the generated targets, using the same task definition provided to the LLM. The annotator received the same input triplet: text, generated target, and generated stance.

We then measured the correlation between the LLM’s coherence judgments and those of the human annotator using the Phi coefficient. The resulting score was 0.6541, indicating a strong positive correlation \citep{njuka2021factors}.

It is important to note that we do not propose a definitive solution or formal methodology for coherence evaluation in stance detection. Rather, we present this as a potential path that can be further explored by the research community. Our experiment suggests that LLMs might serve as reliable judges for assessing the coherence between a generated stance and target within a given text, with judgments that align well with human evaluations. To strengthen the validity of this experiment, future studies could incorporate a wider range of LLMs, multiple runs per sample, and a larger, more diverse set of samples across the three datasets.

\onecolumn

\section{Human Annotation Score Distribution}
\label{app:human_score_dist}
\begin{figure*}[h!]
    \centering
    \includegraphics[width=\textwidth]{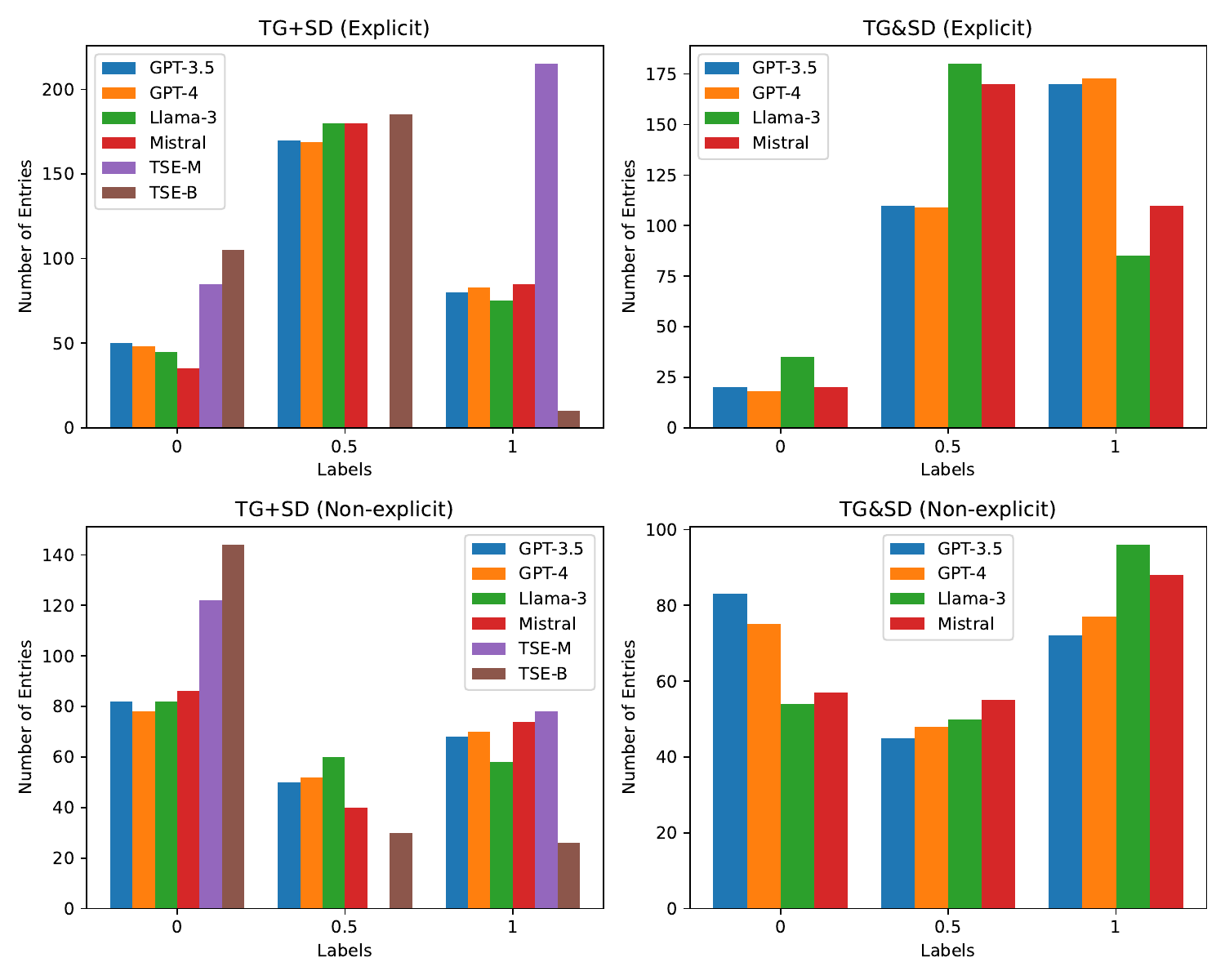}
    \caption{Distribution of human evaluation scores for target relevance assessment (0 – Not Related, 0.5 – Partially or Indirectly Related, 1 – Completely Related). The final score for each sample is determined by the majority vote among the three annotators. In cases of a tie, the average score of 0.5 is assigned. The details on human annotator guideline are provided in Appendix \ref{app:human_ann_guideline}, and the annotator agreement is in Appendix Table \ref{tbl:human_eval_corr}.}
    \label{fig:human_score_dist}
\end{figure*}

After conducting the human annotation for target quality assessment, we plot the score distribution provided by the annotators, as shown in Figure \ref{fig:human_score_dist}. We observe that LLMs generally produce targets that are either partially or fully related to the golden targets, outperforming the TSE models in this regard. The targets generated by TSE-mapped are either identical to the golden targets or entirely unrelated, as \texttt{TSE-mapped} selects targets in its final stage (as described in Section \ref{sec:related_work}) from a pre-defined list. In contrast, the targets generated by \texttt{TSE-BestGen} are often either partially or completely unrelated to the golden targets, indicating lower quality compared to LLMs.

\end{document}